\documentclass[review]{elsarticle}

\usepackage[cmex10]{amsmath}
\usepackage{times}
\usepackage{epsfig}
\usepackage{graphicx}
\usepackage{amssymb}
\usepackage{color}
\usepackage{soul}
\usepackage{enumitem}
\usepackage{todonotes}
\usepackage[normalem]{ulem}
\usepackage[tight,normalsize,sf,SF]{subfigure}
\usepackage{algorithm}
\usepackage{algpseudocode}
\usepackage{hyperref}
\usepackage[title]{appendix}

\graphicspath{{./figs/}}

\begin{document}

\begin{frontmatter}

\title{Continuous Adaptation of Multi-Camera Person Identification Models through Sparse Non-redundant Representative Selection}

\author[myfootnote]{Abir~Das}
\fntext[myfootnote]{The author is currently a postdoctoral researcher at University of Massachusetts, Lowell, MA, 01854}

\author{Rameswar~Panda}

\author{Amit K.~Roy-Chowdhury\corref{mycorrespondingauthor}}
\cortext[mycorrespondingauthor]{Corresponding author}
\ead{amitrc@ee.ucr.com}

\address{University of California, Riverside, CA, 92521}

\begin{abstract}
The problem of image-base person identification/recognition is to provide an identity to the image of an individual based on learned models that describe his/her appearance.
Most traditional person identification systems rely on learning a static model on tediously labeled training data.
Though labeling manually is an indispensable part of a supervised framework, for a large scale identification system labeling huge amount of data is a significant overhead.
For large multi-sensor data as typically encountered in camera networks, labeling a lot of samples does not always mean more information, as redundant images are labeled several times.
In this work, we propose a convex optimization based iterative framework that progressively and judiciously chooses a sparse but informative set of samples for labeling, with minimal overlap with previously labeled images.
We also use a structure preserving sparse reconstruction based classifier to reduce the training burden typically seen in discriminative classifiers.
The two stage approach leads to a novel framework for online update of the classifiers involving only the incorporation of new labeled data rather than any expensive training phase.
We demonstrate the effectiveness of our approach on multi-camera person re-identification datasets, to demonstrate the feasibility of learning online classification models in multi-camera big data applications.
Using three benchmark datasets, we validate our approach and demonstrate that our framework achieves superior performance with significantly less amount of manual labeling.
\end{abstract}

\begin{keyword}
Redundancy reduction, Representative selection, Continuous learning, Person identification/recognition
\end{keyword}

\end{frontmatter}


\section{Introduction}
\label{sec:introduction}

Person identification/recognition across cameras is an important problem in many surveillance and security applications, and requires the ability to handle very large data volumes.
An automated solution of this problem is to use images of persons to provide identities to the images based on learned models that describe his/her appearance.
Most existing solutions depend on learning a static model on tediously labeled training data.
An example of such a task is person re-identification~\cite{Bazzani2012,Kviatkovsky2013,Wang2014,Zhang2013} which is the task of identifying and monitoring people moving across a number of non-overlapping cameras.
Considering the time and labor involved in labeling the training data manually, scalability to large numbers of persons remains an issue.
Also, the learned models being static, cannot adapt to new images that may be available over time.

Some recent semi-supervised~\cite{Zhang2016} and unsupervised methods~\cite{Peng2016,Zhao2013,Liu2012} tried to explore saliency information or weighted features or learning a discriminative null space in a re-identification scenario.
However, none of these works assume a continuous learning setting.
Moreover, it can be seen that unsupervised methods give significantly lower performance compared to supervised methods~\cite{Peng2016,Zhao2013,Liu2012}.
A next alternative is to involve a human in the loop but at the same time efforts should be made to keep the human annotation to a minimum.
Active learning~\cite{Settles2012} is a natural choice for reducing labeling effort by asking for labels only on a few but informative samples (called the active samples), rather than seeking labels either for samples chosen randomly from a set or for the whole set.
In this paper, we explore the question of learning person identification models online in a multi-camera settings with limited labeling effort.
We argue that, in order to truly reduce the labeling cost we need to choose a sparse but informative set of samples to be labeled.
As only a small part of the whole data is annotated, the annotation effort is reduced considerably compared to annotating the whole dataset.
Active learning has been successfully applied to many computer vision problems including tracking~\cite{Vondrick2011}, object detection~\cite{Vijayanarasimhan2011}, image~\cite{Batra2010} and video segmentation~\cite{Fathi2011}, image or scene classification~\cite{Joshi2012,Chakraborty2011,Elhamifar2013} and activity recognition~\cite{Liu2011,Hasan2015,Hasan2014}
However, these methods deal with data coming from single source.
It is not trivial to extend traditional active learning methods for an application scenario where multi-sensor data is involved.
It is a natural challenge to select a few informative samples yet cover as much appearance variation as possible across multiple cameras in such a scenario.
Apart from high cost of labeling the training data, all the data may not also be available at the very outset.
A static pre-trained model can not adapt to the changing dynamics of the incoming data.
In this work, we will address both the above-mentioned scenarios a).  selection of a manageable set of informative samples for labeling and b). doing so in an online manner where all the training data is not available a priori.

For this purpose, we propose an iterative framework which, starting with a pool of unlabeled images, \emph{progressively} and judiciously selects the most informative set of images - termed as the `\textit{representative}' images for labeling with minimal overlap with previously labeled images.
Ideally, a set of representative images are ``representatives'' of a dataset because this set possesses most of the variabilities of the dataset within itself.
On the other hand, without any label information, the representative images are some of the most confusing samples in the whole dataset by the same trait.
Thus annotating such representatives enriches the model by injecting valuable information with a reasonable labeling effort.
We also use a structure preserving sparse reconstruction based classifier to reduce the training burden typically seen in discriminative classifiers.
The use of a sparse classifier enables an \emph{online} update of the identification framework involving only new samples without requiring to train from scratch whenever new batch of data arrives.
This pipeline leads to a framework for online update of the classifiers involving only the incorporation of new labeled data rather than any expensive training phase.
Identifying and eliminating redundant samples is especially important in such an online scenario since reducing redundancy implies more information gain at the cost of less labeling effort.
Thus the proposed work addresses the following question: \textit{Is it possible to select a sparse set of non-redundant training images progressively in an online setting for annotation from multi-sensor data while maintaining good identification performance?}

We demonstrate the effectiveness of our proposed approach on datasets in person re-identification (although our problem setting is different than the traditional re-identification framework).
There are many reasons for it.
Using the re-identification datasets allows us to demonstrate the effectiveness of the online representative selection framework where due to a multi-camera setting, large intra-person variation is prevalent.
Also, these datasets represent uncontrolled settings where we are not dependent on the availability of good quality facial shots.

\subsection{Motivation behind the Proposed Approach}
The representatives or samples chosen iteratively for labeling can have two types of redundancies.
Firstly, in each iteration, the chosen representatives may have many images of the same person.
Secondly, representatives selected in subsequent iterations may also have overlap with the representatives chosen earlier for labeling.
The first type of redundancy is termed as the `intra-iteration redundancy' while the second type is termed as the `inter-iteration redundancy'.
`Intra-iteration redundancy' is restricted by exploiting the fact that redundant samples in any iteration are very close neighbors in the feature space.
Without any feedback about the already chosen representatives, any representative selection strategy may select images of the same person as representatives in subsequent iterations.
Using a similar redundancy reduction strategy of looking for close neighbors as above, we will be able to filter out samples redundant to the already labeled ones in previous iterations.
However, using such a strategy to reduce `inter-iteration redundancy' will prohibit the information gain as images of a person from multiple cameras will be hard to come by.
We tackle this situation by enforcing diversity among the selected representatives as information about the already chosen samples in previous iterations are fed back while choosing subsequent samples to be labeled.
Variabilities are not only caused by the presence of different people but the same person may appear differently in different cameras.
These two different types of variabilities make non-redundant representative selection a challenge in scenarios where there are multiple sources of data as is the case with multi-camera person identification.
The proposed method exploits these variabilities by choosing diverse but small set of representatives from multiple cameras (ref. section~\ref{sec:RedundancyReductionInter}) while discarding similar images of the same person which primarily comes from the same camera (ref. section~\ref{sec:RedundancyReductionIntra}).

Such a representative selection problem is formulated as a \emph{convex optimization} that minimizes the cost of representing an unlabeled pool with a sparse set of representatives as well as one that minimizes the redundancy with the representatives selected earlier.
Experiments on three benchmark datasets show that annotating the small but informative set of representative images reduces the labeling effort considerably, maintaining reasonable identification performance.

Apart from the huge labeling effort, another factor that is a challenge for a scalable solution of the problem is the generally exponential increase of training time with the number of training samples for traditional discriminative classifiers (\textit{e.g.}, SVM or random forest).
These classifiers have to be retrained from scratch after each batch of representative selection and annotation in such repetitive active learning strategy.
The generally super linear time complexity of the traditional discriminative classifiers makes them unsuitable for use in such a scenario.
Though incremental learning based classifiers~\cite{Polikar2001} can update the model without retraining from scratch, their performance is limited by the condition of knowing the number of classes from the start.

Motivated by the recent progress of sparse coding based classifiers~\cite{Deng2012,Wright2009}, we employ a structure preserving sparse dictionary for classification.
Such a classification strategy is helpful as updating the model with newly labeled data means simply adding the new samples with labels without making any changes to the existing dictionary elements made of the already labeled samples.
This model update strategy not only helps in reducing the training time significantly by avoiding the need for retraining but also enables the operation of the framework without assuming any knowledge of the number of classes.
Thus, in summary, the proposed framework uses \emph{two convex optimization based strategies to select a few informative but non-redundant samples for labeling and to update a person identification model online}.

The rest of the paper is organized as follows.
Section~\ref{sec:Relatedworks} discusses the related works and our contributions.
An overview of the proposed approach is given in section~\ref{sec:overview} .
The details about approach, as non-redundant representative selection, and the use of structure preserving sparse coding based classification are described in section~\ref{sec:methodology}.
Experimental results and comparisons are shown in section~\ref{sec:experiments}. Finally, conclusions are drawn in section~\ref{sec:conclusions}.

\section{Related Works and Our Contributions}
\label{sec:Relatedworks}
The proposed method is intricately tied to active learning and representative selection and related to the problem of person re-identification.
This section describes the relevant works these fields have seen in recent years.

{\noindent\ul{\textbf{Active Learning:}}} In an effort to bypass tedious labeling of training data there has been recent interest in `active learning'~\cite{Joshi2012, Vijayanarasimhan2011} where classifiers are trained interactively for incremental update of classification models in presence of streaming video data from a single camera~\cite{Hasan2014}.
Queries are selected for labeling such that enough training samples are procured in minimal effort.
This can be achieved by choosing one sample at a time by maximizing the value of information~\cite{Joshi2012}, reducing the expected error~\cite{Aodha2014} or maximizing both informativeness and representativeness for active sample selection~\cite{Huang2010} prior to retraining a classifier.
On the other hand there have been recent approaches where batches of unlabeled data are selected by exploiting classifier feedback~\cite{Chakraborty2011,Elhamifar2013} or contextual information~\cite{Hasan2015} to maximize informativeness and sample diversity.
To the best of our knowledge, the only work~\cite{Das2015} which takes an active learning approach to multi-camera person identification is using mid level attributes~\cite{Farhadi2009,Parikh2011}.
A `value of information'~\cite{Joshi2012} based strategy is combined with attribute-feedback from the human annotators at the time of annotation.
While the `value of information' strategy helps in reducing the annotation effort by supplying possible sample matches (from the already labeled set) along with selected unlabeled image, the performance of the system depends heavily on the performance of mid level attribute predictors.
Training good attribute predictors calls for a large number of attributes to be annotated which, in turn, calls for large and additional annotation overhead.

{\noindent\ul{\textbf{Representative selection:}}} Most of the applications of representative selection can be found in the fields of video summarization and subset selection.
Historically clustering and vector quantization based methods~\cite{Avila2011,Frey2007,Garcia2012} have dominated these problems, until recently sparse subset selection came into picture.
In~\cite{Cong2012,Elhamifar2012,Elhamifar2012a}, representative selection has been formulated as sparsity regularized linear reconstruction error minimization problem.
The last two works resemble most closely the proposed representative selection framework.
However, without any redundancy restricting condition, the applicability of these frameworks can be limited in a multi-sensor application like person identification as far as reduction of labeling effort is concerned.
A multi-sensor data has its own challenges and redundancy of representatives play a very important role in it.
The Sparse Modeling Representative Selection (SMRS) framework~\cite{Elhamifar2012a} removes redundant frames from an event based summary of videos by considering the proximity of the chosen representative frames in the timeline.
Time information is either unavailable in person re-identification datasets or is unreliable for a such a scenario over a wide space time horizon.
The proposed framework takes care of this issue by splitting the source of redundancy into two parts - one `intra-iteration' and the other `inter-iteration'.
The `intra-iteration' redundancy is reduced by creating a hypergraph between the chosen representatives.
The redundancy among samples chosen in different iterations is reduced by introducing a convex regularization term that minimizes correlation between the new and the previously selected representatives, but at the same time chooses a number of samples representing the data aptly.
This enables the selection of as many difficult examples as possible to improve the identification performance but at the same time avoids labeling a person multiple times unless it is necessary.

{\noindent\ul{\textbf{Person Re-identification:}}}
Our approach being online and adaptive is different from traditional re-identification setting as unlike traditional re-identification scenario, the proposed approach starts with gradually building the gallery set from scratch.
However, due to the challenges of multi-camera re-identification datasets, we are using some benchmark datasets to evaluate our method.
In this section, we will review some recent re-identification works briefly.
Person re-identification approaches can be broadly brought together in three main classes according to the methods followed for solving the problem.
In one class of approaches~\cite{Bazzani2012,Kviatkovsky2013}, camera invariant discriminative signatures have been used to re-identify people in different cameras.
The features used are hand-engineered based on color, shape, texture \textit{etc.}~\cite{Kviatkovsky2013,Bak2012,Liao2015}
Another class~\cite{Liao2015,Li2012a,Kostinger2012,Hirzer2012} has tried to improve the distance measure to better discriminate between different persons by learning the distance metric.
For both the schools of thought deep architecture has proven to be show significant performance boost~\cite{Li2014,Yi2014,Ahmed2015,Ustinova2015}.
However, a common issue in deep architecture based solutions is generating huge amount of labeled training data which has been addressed in the proposed work.
A third class of works tried to explore transformation of features between cameras by learning brightness transfer function~\cite{Javed2008} between appearance features or different variants of it~\cite{Das2014,Prosser2008,Martinel2015}.
Some recent unsupervised methods~\cite{Zhao2013,Liu2012} tried to explore saliency information or weighted features towards re-identifying people across cameras.
However, none of these methods consider an interactive framework that selects the most informative set of representatives for manual labeling, thus reducing the effort of the human.
For a thorough review of the person re-identification literature, interested readers are directed to the review paper~\cite{Vezzani2013} where a multidimensional taxonomy and categorization of the person re-identification algorithms can be obtained.

{\noindent\underline{\textbf{Contributions of the paper:}}} To summarize, the contributions of the proposed approach to the problem of person identification are the followings.
\begin{itemize}[label={--},noitemsep,leftmargin=*,topsep=0pt,partopsep=0pt]
  \item Large-scale person identification has been formulated as an multi-sensor active learning system with an eye to reduce the huge annotation effort arising out of sheer volume of the training data.
  \item A sparsity regularized convex optimization framework has been proposed to deal with non redundant representative selection, crucial to reducing the labeling effort in a multi-camera setting through iterative active learning framework.
  The framework can, in general, be used for other multi-sensor active learning frameworks, \textit{e.g.}, activity recognition where data from multiple sensors may arrive in batches.
  \item Through a set of experiments, we show that the proposed framework can give significantly higher identification accuracy for a fixed labeling effort.
  New standardizations of person identification experiments have been introduced specifically geared towards analyzing this aspect.
\end{itemize}

\section{Overview of the Proposed Approach}
\label{sec:overview}

\begin{figure*}[t]
\centering
\includegraphics[width=1\linewidth]{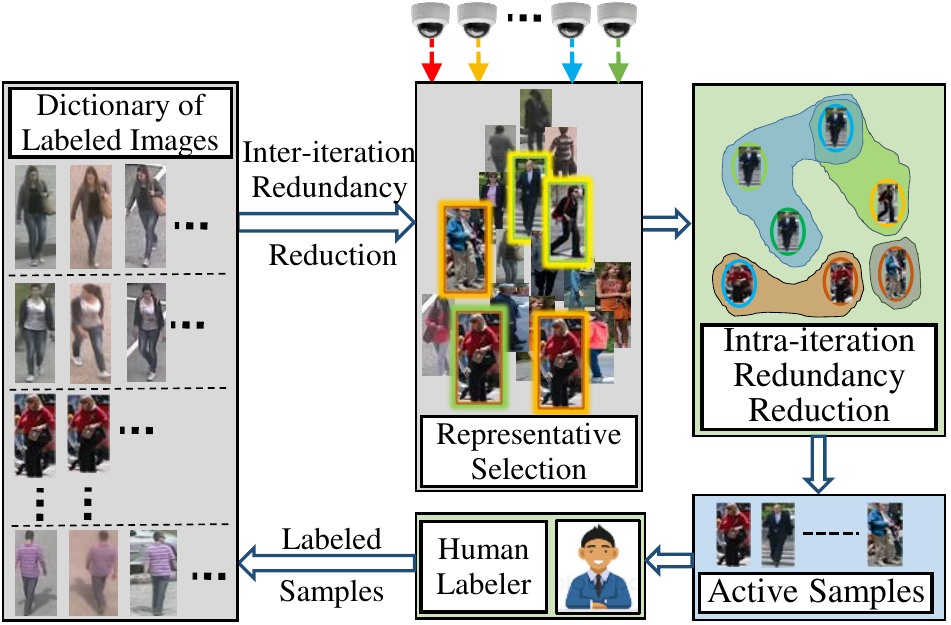}
\caption{System Overview.
The `Representative Selection' module (shown in the middle of the top row) receives unlabeled images of persons from multiple sensors and selects a few informative representatives from them.
Next, redundant images inside the samples chosen in this step are removed inside the `intra-iteration redundancy reduction' module.
Now the active samples or the representatives filtered by this `intra-iteration redundancy reduction' module are presented to the human annotators seeking for labels.
The labeled samples form a dictionary which is fed to the representative selection framework so that in the next iteration those representatives from the unlabeled pool are chosen which are maximally non-redundant with the labeled images in the dictionary.
This cycle goes on as new images come from the streaming videos.}
\label{fig:system}
\end{figure*}

The overall scheme of the proposed approach is shown in Fig.~\ref{fig:system}.
Given incoming streaming videos and detected person images, the proposed framework iteratively choses small sets of informative images to be labeled by human annotators.
These informative images, called the `active samples' are chosen starting with completely unlabeled pool of detections.
The human annotators gives identities (labels) to the active samples by comparing with a gallery of already labeled samples (denoted here as `Dictionary of labeled images').
Initially the labeled dictionary is empty and is incrementally built as more and more data becomes available.
Such a framework makes the proposed approach suitable for an online application scenario where all the data is not available a-priori and the data comes in batches.
At each iteration, new sets of active samples are chosen by the proposed approach (shown as the `Representation Selection' block in Fig.~\ref{fig:system}) from the unlabeled pool to be sen to the human annotators.
It should be noted that an iteration of `active sample' selection does not depend on the arrival of the batch of input data.

Next, the redundant images from the chosen representatives are eliminated by forming a hypergraph between the representative samples and choosing one image per hyperedge of the hypergraph.
This is shown in the block `Intra-iteration Redundancy Reduction' in Fig.~\ref{fig:system}.
Overlapping hyperedges in such a hypergraph, contain images of very similar looking persons.
So images which are mutually exclusive as well as the images common between the overlapping hyperedges are of utmost importance as labeling them helps in disambiguation between difficult (similar looking) persons.
Now the active samples filtered by the intra-iteration redundancy reduction are presented to the human annotators for labels.
These labeled samples are stacked in a dictionary which has two usages.
As shown in Fig.~\ref{fig:system}, these images are fed to the representative selection framework along with the rest of the unlabeled images in the next iteration.
The resulting convex optimization, now, minimizes the correlation between the already labeled samples and the unlabeled samples along with the reconstruction error term.
This cycle goes on until a predefined number of samples are annotated.

The second usage of the labeled dictionary (not shown in the figure) is for identification of unknown samples in a Sparse Reconstruction based Classification (SRC)~\cite{Deng2012,Wright2009} framework where the set of labeled samples work as the dictionary for the SRC.
Assuming that the intra-person variations of one subject can be approximated by a sparse linear combination of its labeled images, SRC finds the sparse representation of each unlabeled test image in terms of the labeled images in the dictionary.
The sparse coefficients when grouped by the labels of the dictionary elements and subsequently normalized give the probability distribution of the unlabeled test images to be one of the labeled subjects.

\section{Methodology}
\label{sec:methodology}
In this section, our proposed framework is discussed in details.
First we describe the notations that would be used throughout the rest of the section before providing the problem statement and basic formulation for representative selection.
Next we will discuss how two different types of redundancies as described in section~\ref{sec:introduction} are restricted while choosing the small but informative set of representatives.
We will discuss about the intra-iteration redundancy reduction first and then the inter-iteration redundancy reduction will be detailed.
Finally, in this section, we will see the optimization strategy to solve the problem.

\subsection{Problem Statement and Notation}
\label{sec:PrblmStmnt}
We use boldface uppercase letters (\textit{e.g.}, $\mathbf{X}$) to denote matrices.
A superscript(\textit{e.g.}, $\mathbf{x}^{(i)}$)/subscript(\textit{e.g.}, $\mathbf{x}_i$) associated with a boldface lowercase letter will denote the corresponding row/column of the matrix.
A boldface lowercase letter will denote a column vector, unless otherwise specified.
The $ij^{th}$ element of the matrix $\mathbf{X}$ will be denoted as $\mathbf{X}_{ij}$.
$tr(.)$ denotes the trace operator.
$diag(.)$ denotes the diagonal operator which extracts the main diagonal of a matrix.

We start with a large pool of unlabeled images containing instances of persons from different cameras.
This defines the input to the framework.
Let at a certain iteration, the features from $n$ unlabeled images be arranged as columns of the matrix $\mathbf{Z} = [\mathbf{z}_1, \mathbf{z}_2, \cdots, \mathbf{z}_n]$, where $\mathbf{z}_i \in \mathbb{R}^d$ denotes the $d$ dimensional feature vector from the $i^{th}$ image.
We aim to select a sparse set of columns (say, $k$ number of columns where $k\ll n$) which represents the whole collection $\mathbf{Z}$.
The corresponding images are the output of the non-redundant representative selection framework which is labeled by the human annotators.
The different variables used to describe the proposed approach are summarized in Table~\ref{tab:Notation}.

\begin{table}[]
	\centering
	\caption{Table with definitions and descriptions of different variables used in the proposed work.}
	\label{tab:Notation}
	\begin{tabular}{|p{1.8cm}|p{9.5cm}|}
	\hline
	\textbf{Variables} &  \textbf{Descriptions}\\ \hline
		$\mathbf{Z} \in \mathbb{R}^{d\times n}$ &  The set of $n$ unlabeled images where each column is a $d$ dimensional vector.\\ \hline
		$\mathbf{\widehat{Z}}_0 \in \mathbb{R}^{d\times n_0} $ & The set of $n_0$ labeled images where each column is again a $d$ dimensional vector. This is initially a null matrix. This also serves as the labeled dictionary.\\ \hline
		$\mathbf{X} \in \mathbb{R}^{n\times n}$ &  Unknown selection matrix whose nonzero rows give the indices of the selected representatives in terms of the columns of $\mathbf{Z}$.\\ \hline
		$\lambda_1, \lambda_2$ & Regularization parameters for the inter-iteration redundancy reduction and the sparsity controlling terms respectively (ref. eqn.~(\ref{eq:rep_selection})).\\ \hline
		$\mathbf{Y}$ &  The set of unlabeled test images where each column represents the feature vector from each test image.\\ \hline
		$\mathbf{\widehat{Y}}_0$ &  The set of annotated images where each column represents the feature vector from each annotated image.\\ \hline
		$\mathbf{C}$ &  The sparse representation of the set of unlabeled test images $\mathbf{Y}$ where each column is $n_0$ dimensional sparse representation of the corresponding columns of $\mathbf{Y}$.\\ \hline
		$\mathbf{L}$ &  The graph Laplacian Matrix~\cite{Luxburg2007} obtained from a k-nearest neighbor graph of similarities calculated between the columns of $\mathbf{Y}$.\\ \hline
		$\alpha, \beta$ & Regularization parameters for the sparse coefficient matrix and the structure preserving term respectively (ref. eqn.~(\ref{eq:SRC_LSc})).\\ \hline
		$g(\mathbf{X})$ & The sum of the reconstruction error and the inter-iteration redundancy reduction terms $(||\mathbf{Z} - \mathbf{ZX}||_F^2 + \lambda_1 ||\mathbf{\widehat{Z}}_0^T\mathbf{ZX}||_F^2)$ as defined in eqn.~(\ref{eq:g_X}). \\ \hline
		$\nabla g(\mathbf{X})$, $L_g$ & The gradient of $g(\mathbf{X})$ and the Lipschitz constant of the gradient respectively. \\ \hline
		$p(\mathbf{C})$ & The sum of the reconstruction error and the structure preserving terms $(||\mathbf{Y} - \mathbf{\widehat{Y}}_0\mathbf{C}||_F^2 + \beta tr(\mathbf{CLC}^T))$ in eqn.~(\ref{eq:SRC_LSc}) \\ \hline
		$\nabla p(\mathbf{C})$, $L_p$ & The gradient of $p(\mathbf{C})$ and the Lipschitz constant of the gradient respectively. \\ \hline
	\end{tabular}
\end{table}

\subsection{Basic Formulation}
\label{sec:subBasicFormul}
Finding compact dictionaries~\cite{Aharon2006,Lee2006,Zheng2011} has been studied as a way to represent data.
Such approaches find the dictionary elements by searching for a set of basis vectors which expresses the data in terms of sufficiently sparse coefficient vectors with respect to the basis vectors.
However, the basis vectors \textit{i.e.}, the elements of the dictionary hardly coincides with the original data and thus do not serve as representatives selected from the data itself.
To find representatives from the data itself we use the following basic formulation which defines a combinatorial optimization.
This is subsequently relaxed later after it is suitably constrained by the redundancy restriction term.
\begin{gather}
\label{eq:basic_formulation_0}
\underset{\mathbf{D,U}}{\operatorname{min}}\, ||\mathbf{Z} - \mathbf{ZDU}||_F^2\\
\text{s.t. $\mathbf{D}$ is $n\times n$ diagonal matrix with } diag(\mathbf{D}) \in \{0,1\}^{n\times 1}, \nonumber\\
||diag(\mathbf{D})||_0 = k, \text{ and } \mathbf{U} \in \mathbb{R}^{n\times n} \nonumber
\end{gather}

Here, $||.||_F$ denotes the Frobenius norm of a matrix and $||.||_0$ denotes the zero norm of a vector.
$\mathbf{D}$ is a $n\times n$ diagonal matrix with only $0$ or $1$ in its diagonal.
The $\ell_0$ norm of the diagonal of such a matrix being $k$ signifies that only $k$ of the $n$ diagonal elements are $1$, rest are $0$.
Such a binary diagonal matrix when multiplied with $\mathbf{Z}$, selects only $k$ columns out of the $n$ columns of $\mathbf{Z}$.
$\mathbf{U}$ is a full real matrix with $n$ rows and $n$ columns.
While post multiplication of $\mathbf{Z}$ by $\mathbf{D}$ selects $k$ columns of $\mathbf{Z}$, further post multiplication of the product by the full real matrix $\mathbf{U}$ linearly combines the selected $k$ columns of $\mathbf{Z}$ so that the resultant matrix $\mathbf{ZDU}$ is as close as possible to the original matrix $\mathbf{Z}$.
Here, both $\mathbf{D}$ and $\mathbf{U}$ are unknown.
Lets denote the product of the two unknowns $\mathbf{DU}$ as $\mathbf{X}$.
Since $\mathbf{D}$ is a diagonal matrix with only $k$ 1's and $n-k$ 0's in its diagonal and $\mathbf{U}$, in general, is a full matrix, the product $\mathbf{X}$ will be matrix whose $k$ rows will be non-zero while $n-k$ rows will be all zeros.
The indices of the non-zero rows of $\mathbf{X}$ correspond to the indices of 1's in $\mathbf{D}$ which, in turn, correspond to the selected columns (as representatives) of $\mathbf{Z}$.
All these characteristics of such a matrix $\mathbf{X}$ can be conveniently and succinctly expressed in terms of $\ell_{2,0}$ matrix norm.
$\ell_{2,0}$ matrix norm is the number of non-zero rows of a matrix.
So all the constraints in eqn.~(\ref{eq:basic_formulation_0}) can be written as $||\mathbf{X}||_{2,0} = k$ where $\mathbf{X} \in \mathbb{R}^{n\times n}$.
Thus, changing the constraints in terms of $\ell_{2,0}$ norm, we write the basic formulation in eqn.~(\ref{eq:basic_formulation_0}) as follows,

\begin{gather}
\label{eq:basic_formulation}
\underset{\mathbf{X}}{\operatorname{min}}\, ||\mathbf{Z} - \mathbf{ZX}||_F^2\\
\text{s.t. } \mathbf{X} \in \mathbb{R}^{n\times n}, ||\mathbf{X}||_{2,0} = k \nonumber
\end{gather}

The indices of the non-zero rows of $\mathbf{X}$ will give the column indices of the selected representatives from $\mathbf{Z}$.

\subsection{Reduction of Redundancy among Chosen Representatives}
\label{sec:RedundancyReduction}
The representatives selected in this way will suffer from the fact that they will contain images of the same person which, in turn, will increase the human effort of labeling.
At the same time, the system will not be adaptive to possible new input data as it will have less access to new and diverse training data due to the presence of similar (redundant) samples.
As discussed in section~\ref{sec:introduction}, images can be redundant among several iterations of chosen representatives as data continues to come and representatives are chosen in several batches (inter-iteration redundancy).
On the other hand, redundant images within an iteration are termed as intra-iteration redundant images and these also need to be as less as possible to reduce the annotation effort.
First, we briefly discuss about reducing the intra-iteration redundancy and then we will see how the proposed method addresses the inter-iteration redundancy.

\subsubsection{Reduction of Intra-iteration Redundancy}
\label{sec:RedundancyReductionIntra}
We have seen that $\mathbf{ZX}$ gives the reconstructed pool of unlabeled images as a linear combination of the selected representatives where the selected representatives of $\mathbf{Z}$ are given by the indices of the non-zero rows of $\mathbf{X}$.
There can be repetitive selection of images resulting in intra-iteration redundancy between the selected representatives.
This is reduced by forming a hypergraph among the representatives selected by solving eqn.~(\ref{eq:rep_selection}).
hypergraphs~\cite{Berge1984} are a generalization of graphs where one edge can be connected to any number of nodes.
Such an edge is named as a hyperedge which links a subset of nodes instead of two nodes only as in ordinary graphs.
In this sense, an ordinary graph is a special kind of hypergraph.
After each iteration, such a hypergraph is formulated where the nodes of the hypergraph are the chosen active samples in that particular iteration.
The hyperedges, created in the feature space itself, contains the redundant samples.
From the `$k\times k$' feature similarity matrix, a `$k\times k$' \emph{adjacency matrix} is created using a high threshold of feature similarity.
The adjacency matrix subsequently gives the`$m\times k$' \emph{incidence matrix} where `$m$' is the number of hyperedges.
Note that such a graph based clustering has major advantage over popular and simple clustering methods \textit{e.g.}, k-means as the success of k-means depends largely on the judicious choice of k.
Hypergraph based redundancy reduction depends on the threshold of the similarity score.
The role of the threshold is to create hyperedges so that each hyperedge contains highly similar images.
Since, only one image per hyperedge is given a label (by the annotator), the hyperedges must contain similar images with high confidence so that otherwise informative (i.e. non-redundant) samples do not get discarded.
This is ensured by choosing a high value of this threshold, leading to high precision at the cost of possibly low recall value in finding similar representatives inside an iteration.
As all the images in each group of redundant samples are given a single identity, the use of such high threshold prevents the model to get updated with wrong labels.

\subsubsection{Reduction of Inter-iteration Redundancy}
\label{sec:RedundancyReductionInter}
Though the formulation in eqn.~(\ref{eq:basic_formulation}) selects a sparse set of representative images for labeling, but it is less effective in dealing with the `inter-iteration' redundancy.
Let us denote the set of selected representatives till a certain iteration by $\mathbf{\widehat{Z}}_0$ which is a matrix of dimension $d\times n_0$ containing the features from the already selected $n_0$ images.
Now, $\mathbf{Z}$ contains the features from the rest of the unlabeled images.
For convenience let us write the reconstructed features from this rest of the unlabeled images $\mathbf{ZX}$ as $\mathbf{\widehat{Z}}$.
Without any loss of generality, let us assume that all the features are made zero mean.
In that case, we show below that $||\mathbf{\widehat{Z}}_0^T\mathbf{\widehat{Z}}||_F^2$ expresses the correlation between the already selected images and the rest.
This is because,
\begin{flalign}
\label{eq:trace_corr_term}
||\mathbf{\widehat{Z}}_0^T\mathbf{\widehat{Z}}||_F^2 &= \sum_{i=1}^{n_0} \sum_{j=1}^{n-n_0} \big[\big(\mathbf{\widehat{Z}}_0^T\mathbf{\widehat{Z}}\big)_{ij} \big]^2 = \sum_{i=1}^{n_0} \sum_{j=1}^{n-n_0} \big[(\mathbf{\widehat{z}}_0)_i^T\mathbf{\widehat{z}}_j \big]^2 \nonumber\\
&= \sum_{i=1}^{n_0} \sum_{j=1}^{n-n_0} d^2 \sigma_i^2 \sigma_j^2 \rho_{ij}^2
\end{flalign}
where $\sigma_i$ denotes the standard deviation of the features of the $i^{th}$ image in $\mathbf{\widehat{Z}}_0$ and likewise $\sigma_j$ denotes the standard deviation for the $j^{th}$ image in $\mathbf{\widehat{Z}}$.
$\rho_{ij}$ denotes the correlation coefficient between the features of the $i^{th}$ image in $\mathbf{\widehat{Z}}_0$ and the $j^{th}$ image in $\mathbf{\widehat{Z}}$.
The last line in eqn.~(\ref{eq:trace_corr_term}) is due to the fact that all the columns of both $\mathbf{\widehat{Z}}_0$ and $\mathbf{\widehat{Z}}$ have been converted to zero means.
From this, it can be seen that minimizing $||\mathbf{\widehat{Z}}_0^T\mathbf{\widehat{Z}}||_F^2$ prefers to select the columns of $\mathbf{Z}$ which are less correlated to the images in $\mathbf{\widehat{Z}}_0$.
So, adding $||\mathbf{\widehat{Z}}_0^T\mathbf{\widehat{Z}}||_F^2$ (\textit{i.e.}, $||\mathbf{\widehat{Z}}_0^T\mathbf{ZX}||_F^2$) as a regularizer to eqn.~(\ref{eq:basic_formulation}) makes sure that a sparse set of images non-redundant with previously selected representatives are obtained.
Using a regularization parameter $\lambda_1$ the problem can now be written as,
\begin{gather}
\label{eq:non_relaxed_formulation}
\underset{\mathbf{X}}{\operatorname{min}}\, ||\mathbf{Z} - \mathbf{ZX}||_F^2 + \lambda_1 ||\mathbf{\widehat{Z}}_0^T\mathbf{ZX}||_F^2\\
\text{s.t. } \mathbf{X} \in \mathbb{R}^{n\times n}, ||\mathbf{X}||_{2,0} = k \nonumber
\end{gather}

In Eqn.~(\ref{eq:non_relaxed_formulation}), the first term of the cost function minimizes the reconstruction error of the feature from each image when the reconstruction is done as a linear combination of features from the selected representative images.
The second term minimizes the correlation between the selected representatives and the previously selected ones.
The constraint on $\ell_{2,0}$ norm of $||\mathbf{X}||$ implies that only $k$ rows of it will be non-zero.
In a nutshell, the nonzero rows of $\mathbf{X}$ correspond to those columns of $\mathbf{Z}$ which represent the whole unlabeled pool $\mathbf{Z}$ with minimal overlap with the previously selected representatives.

{\noindent\ul{\textbf{Relaxation of the Constraints:}}}
Eqn.~(\ref{eq:non_relaxed_formulation}) is NP-hard and can be highly computationally expensive even for moderate values of $k$ and $n$.
We need to relax the optimization problem in eqn.~(\ref{eq:non_relaxed_formulation}) to make it a convex optimization problem as the $\ell_{2,0}$ norm is non-convex.
Following the common strategy of 1-norm relaxation for 0-norms, we employ $||.||_{2,1}$ norm in place of the $\ell_{2,0}$ norm and reformulate the problem as,
\begin{equation}
\begin{gathered}
\label{eq:relaxed_formulation_21}
\underset{\mathbf{X}\in \mathbb{R}^{n\times n}}{\operatorname{min}}\, ||\mathbf{Z} - \mathbf{ZX}||_F^2 + \lambda_1 ||\mathbf{\widehat{Z}}_0^T\mathbf{ZX}||_F^2 \\
\text{ s.t. } ||\mathbf{X}||_{2,1} \leq k
\end{gathered}
\end{equation}

{\noindent\ul{\textbf{Overall Optimization Problem:}}}
Using Lagrange multipliers, the overall optimization problem from eqn.~(\ref{eq:relaxed_formulation_21}) can be written as,
\begin{equation}
\label{eq:rep_selection}
\underset{\mathbf{X}}{\operatorname{min}}\, ||\mathbf{Z} - \mathbf{ZX}||_F^2 + \lambda_1 ||\mathbf{\widehat{Z}}_0^T\mathbf{ZX}||_F^2 + \lambda_2 ||\mathbf{X}||_{2,1}
\end{equation}
where, $\lambda_1$ and $\lambda_2$ are the two regularization parameters.
The inputs to the optimization problem are the unlabeled images $\mathbf{Z}$ and the labeled images $\mathbf{\widehat{Z}}_0$ while the output is the selection matrix $\mathbf{X}$ whose non-zero row indices provide the representative images to be labeled.
After labeling, the annotated samples are inducted into the dictionary as dictionary elements and the identification probability of the test images are obtained by finding sparse representations of the test samples with respect to the dictionary according to the formulation described next.

The conversion of the constrained optimization problem to the corresponding unconstrained problem as in eqn.~(\ref{eq:rep_selection}), employing Lagrange multipliers brings in independence from $k$.
That is, the number of non-zero rows may not be exactly $k$.
Following standard practice in literature~\cite{Cong2012,Elhamifar2012a} we choose the top $k$ rows in terms of their 2 norms when the number of non-zero rows of $\mathbf{X}$ is greater than $k$.
For the case when this number is less than $k$, only all the non-zero rows are taken.
The representatives are chosen from the corresponding columns of $\mathbf{Z}$.

\subsection{Classification and Online Update}
\label{sec:Sparse_Dict}
The chosen samples are annotated by the human annotators and the annotated samples form the dictionary elements.
The dictionary is used to find the probability of the test samples via finding the sparse representations of the test samples.
The active samples are obtained by solving eqn.~(\ref{eq:rep_selection}).
Once the active samples are annotated, the sparse classification dictionary can be formed of any features extracted of these annotated samples.

Using $\mathbf{\widehat{Y}}_0$ and $\mathbf{Y}$ to denote features extracted from the annotated representatives and the test samples, the sparse representation of $\mathbf{Y}$ can be found by minimizing the following.
\begin{equation}
\label{eq:SRC}
\underset{\mathbf{C}}{\operatorname{min}}\, ||\mathbf{Y} - \mathbf{\widehat{Y}}_0\mathbf{C}||_F^2 + \alpha ||\mathbf{C}||_1
\end{equation}
Ideally a test image is reconstructed from a linear combination of labled samples from the same class as that of the test sample.
The sparsity condition makes sure that training samples from other classes appear as infrequently as possible in the reconstruction of the test image.
Seeking the sparsest representation, therefore, discriminates between the various classes of test samples and the sparse coefficients (when normalized) provide the probability of the test sample to belong to that class.
However, the overcomplete nature of the dictionary can give rise to loss in structure of the data.
Similar features may be encoded by different sparse codes giving rise to entirely different probability distribution for samples of same class~\cite{Qiu2011}.

To increase the robustness of a sparse code based classifier, graph Laplacian has been used~\cite{Zheng2011,Gao2013}.
After incorporating the structure preserving regularizer in eqn.~(\ref{eq:SRC}), the sparse classifier can be written as,
\begin{equation}
\label{eq:SRC_LSc}
\underset{\mathbf{C}}{\operatorname{min}}\, ||\mathbf{Y} - \mathbf{\widehat{Y}}_0\mathbf{C}||_F^2 + \alpha ||\mathbf{C}||_1 + \beta\, tr(\mathbf{CLC}^T)
\end{equation}
where $\mathbf{L}$ is the graph Laplacian~\cite{Luxburg2007} obtained from a k-nearest neighbor graph of similarities calculated between the columns of $\mathbf{Y}$.
Here, $\alpha$ and $\beta$ are regularization parameters.
Using such a sparsity based strategy we are able to update the classifier online simply by incorporating the labeled images in any iteration to the already existing dictionary.
Unlike the discriminative classifiers this involves no expensive training phase and thus online update of the classification model is not an overhead with large number of classes.

\subsection{Optimization}
\label{sec:Optimization}
Here we state the optimization strategy to solve the two convex optimization problems (eqns.~(\ref{eq:rep_selection}) and~(\ref{eq:SRC_LSc})).
Both the equations involve convex but non-smooth terms which require special attention.
Proximal methods are specifically tailored towards it.
These methods have drawn increasing attention in the machine learning community because of their fast convergence rates.
They find the minimum of a cost function of the form $g(\mathbf{X}) + h(\mathbf{X})$ where $g$ is convex, differentiable but $h$ is closed, convex and non-smooth.
We use fast proximal algorithm, FISTA~\cite{Beck2009} which maintains two variables in each iteration and combines them to find the solution.
New value of the variable, in each iteration is computed by computing the proximal operator of $h$ on a function of the gradient of $g$. (ref eqn.~(\ref{eq:FISTA_Basic})).
The proximal operator of $h(\mathbf{X})$, denoted as $\text{Prox}_h(\mathbf{X})$ is computed as,
\begin{equation}
\label{eq:prox_op}
\text{Prox}_h(\mathbf{X}) = \underset{\mathbf{U}}{\operatorname{argmin}}\,\, \big(h(\mathbf{U}) + \frac{1}{2} ||\mathbf{U} - \mathbf{X}||^2 \big)
\end{equation}
where, the domain of $\mathbf{U}$ is the set of real matrices with same dimension as $\mathbf{X}$.
The FISTA algorithm can be summarized by the following two steps after choosing any initial $\mathbf{X}^{(0)} = \mathbf{X}^{(-1)}$ (the superscripts, here, denote the iteration number of FISTA).
\begin{equation}
\begin{gathered}
\label{eq:FISTA_Basic}
\text{\textbf{Step I: }} \mathbf{Y} = \mathbf{X}^{(k-1)} + \frac{k-2}{k-1}\big( \mathbf{X}^{(k-1)} - \mathbf{X}^{(k-2)} \big) \\
\text{\textbf{Step II: }} \mathbf{X}^{(k)} = \text{Prox}_{t_k h}(\mathbf{Y} - t_k \nabla g(\mathbf{Y}))
\end{gathered}
\end{equation}
where, $k$ is the iteration index (of FISTA) and $t_k$ is the step size parameter.

{\noindent\ul{\textbf{Gradients and Lipscschitz's constants:}}}
In eqn.~(\ref{eq:rep_selection}), the reconstruction error and the inter-iteration redundancy reduction terms are convex, smooth, differentiable functions with Lipschitz continuous gradients.
Let us denote the sum of these two terms as $g(\mathbf{X})$ \textit{i.e.},
\begin{equation}
\label{eq:g_X}
g(\mathbf{X}) = ||\mathbf{Z} - \mathbf{ZX}||_F^2 + \lambda_1 ||\mathbf{\widehat{Z}}_0^T\mathbf{ZX}||_F^2
\end{equation}
The gradient $\nabla g(\mathbf{X})$ and the Lipschitz constant $L_g$ of the gradient are given by,
\begin{equation}
\begin{gathered}
\label{eq:GradLipRepSel}
\nabla g(\mathbf{X}) = 2\big(-\mathbf{Z}^T\mathbf{Z} + \mathbf{Z}^T\mathbf{ZX} + \lambda_1 \mathbf{Z}^T\mathbf{\widehat{Z}}_0 \mathbf{\widehat{Z}}_0^T\mathbf{ZX} \big)\\
L_g = 2\big( ||\mathbf{Z}^T\mathbf{Z}||_F^2 + \lambda_1 ||\mathbf{Z}^T\mathbf{\widehat{Z}}_0 \mathbf{\widehat{Z}}_0^T\mathbf{Z}||_F^2 \big)
\end{gathered}
\end{equation}
Similarly, the reconstruction error and the structure preserving terms in eqn.~(\ref{eq:SRC_LSc}), are convex, smooth and differentiable functions of $\mathbf{C}$.
Denoting $||\mathbf{Y} - \mathbf{\widehat{Y}}_0\mathbf{C}||_F^2 + \beta tr(\mathbf{CLC}^T)$ as $p(\mathbf{C})$, the gradient $\nabla p(\mathbf{C})$ and the Lipschitz constant $L_p$ of the gradient are given by,
\begin{equation}
\begin{gathered}
\label{eq:GradLipSparseRecon}
\nabla p(\mathbf{C}) = 2\big(- \mathbf{\widehat{Y}}_0^T\mathbf{Y} + \mathbf{\widehat{Y}}_0^T\mathbf{\widehat{Y}}_0\mathbf{C} + \beta\mathbf{CL}\big)\\
L_p = 2\big(||\mathbf{\widehat{Y}}_0^T\mathbf{\widehat{Y}}_0||_F^2 + \beta||\mathbf{L}||_F^2\big)
\end{gathered}
\end{equation}

\begin{algorithm}
\caption{Overall Framework}\label{algo:Overall}
\begin{algorithmic}
\State {\bf \ul{Active Training:}}
\State {\bf Input:} \ul{Data:} Unlabeled images $\mathbf{Z}$,~~ \ul{Parameters:} $\lambda_1, \lambda_2, T$ (\# of iterations)
\State {\bf Output:} Representatives for labeling $\mathbf{\widehat{Z}}_0$
\State $\mathbf{\widehat{Z}}_0 \gets \phi$ (null set), 
\For {$ i \leftarrow 1$ to $T$} 
\State {$\mathbf{X} \gets$ solution of eqn.~(\ref{eq:rep_selection}) by FISTA (eqn.~\ref{eq:FISTA_Basic} and \ref{eq:prox_h}) using gradient $\nabla g(\mathbf{X})$ and Lipschitz constant $L_g$ (eqn.~\ref{eq:GradLipRepSel})}
\State {$\mathbf{Z}_s \gets$ columns of $\mathbf{Z}$ corresponding to non-zero rows of $\mathbf{X}$}
\State {$\mathbf{\widehat{Z}}_0 \gets \mathbf{\widehat{Z}}_0 \cup \mathbf{Z}_s, \, \mathbf{Z} \gets \mathbf{Z} \setminus \mathbf{Z}_s$}
\EndFor
\\
\State {\bf \ul{Test:}}
\State {\bf Input:} \ul{Data:} $\mathbf{Y}, \mathbf{\widehat{Y}}_0$,~~ \ul{Parameters:} $\alpha, \beta$
\State {\bf Output:} Sparse coefficient matrix $C$
\State {Compute $\mathbf{L}$} from $\mathbf{Y}$ (ref. section \ref{sec:RedundancyReductionIntra})
\State {$\mathbf{C} \gets$ solution of eqn.~(\ref{eq:SRC_LSc}) by FISTA (eqn.~\ref{eq:FISTA_Basic} and \ref{eq:prox_q}) using gradient $\nabla p(\mathbf{C})$ and Lipschitz constant $L_p$ (eqn.~\ref{eq:GradLipSparseRecon})}
\end{algorithmic}
\end{algorithm}

{\noindent\ul{\textbf{Proximal operators:}}}
The sparsity inducing $\ell_{2,1}$ norm (in eqn.~\ref{eq:rep_selection}) and the $\ell_1$ norm (in eqn.~\ref{eq:SRC_LSc}) both are convex but non-smooth functions of their respective variables.
Let us denote the non-smooth terms as $h(\mathbf{X})$ and $q(\mathbf{C})$ respectively, \textit{i.e.}, $\lambda_2 ||\mathbf{X}||_{2,1} = h(\mathbf{X})$ and $\alpha ||\mathbf{C}||_1 = q(\mathbf{C})$.
The corresponding proximal operators for these two non-smooth functions are given by,
\begin{align}
\text{Prox}_h(\mathbf{X}) = \big(1- \frac{\lambda_2}{||\mathbf{X}^{(i)}||_2} \big)_+ \mathbf{X}^{(i)}\label{eq:prox_h}\\
\text{Prox}_q(\mathbf{C}) = \big(1- \frac{\alpha}{|\mathbf{C}_{ij}|} \big)_+ \mathbf{C}_{ij} \label{eq:prox_q}
\end{align}
where $i$ and $j$ denote the row and column numbers with $(x)_+ \triangleq \text{max}(x,0)$.
Taking a fixed step size $t_k$ equal to the inverse of the respective Lipschitz constants, the convergence rate of FISTA is proportional to $\frac{1}{k^2}$, in contrast to $\frac{1}{\sqrt{k}}$ in subgradient based methods where $k$ denotes the iteration number.
The overall iterative framework towards online and interactive person identification using the gradients, Lipschitz constants and the proximal operators is presented in algorithm~\ref{algo:Overall}.

\section{Experiments}
\label{sec:experiments}

This section gives details about the different experimental scenarios with specific objectives and experimental results validating our proposed approach.
We start with a description of the feature set followed by describing the experimental scenarios and then discussion of the results.

{\noindent\textbf{Feature Extraction:}} 
Mean color feature (HSV) is used following the scheme in~\cite{Hirzer2012}.
Since the images are from different cameras, the features can vary a lot due to the changes of several factors including but not limited to scale, illumination, depth \textit{etc}.
However, for a single person as the features are coming from the same person irrespective of the camera, it is reasonable to assume that the features from the same person are close to each other in some underlying joint manifold.
This directed us to find a low dimensional manifold out of the features from the unlabeled pool of images.
On the other hand, sparse representative selection methods have a tendency to select redundant representatives since they fail to capture the locality and correlations present in the original data~\cite{Elhamifar2012a,Chennubhotla2001}.
To achieve this, we use a t-SNE~\cite{Maaten2008a} dimensionality reduction technique to preserve the data correlations~\cite{Chennubhotla2001}, before a sparse representative selection method is applied.
t-SNE projects the appearance features extracted from the unlabeled pool of images to a low dimensional manifold prior to employing eqn. (\ref{eq:rep_selection}).
For experimentation purposes, we apply t-SNE on the whole unlabeled data at once to project it into a low dimensional space before the active sample selection.
However, out-of-sample-extension ideas~\cite{Strange2011} can also be used to find the low dimensional coordinate of the new incoming samples in an online manner.
It should be noted that use of t-SNE is completely independent of the SRC as t-SNE mapped features are only used for active sample selection (ref. eqn. (\ref{eq:rep_selection})).
Once the representatives are selected and annotated, the SRC takes the original features of the active samples as to form the dictionary.

{\textbf{Experiment Design:}} The experiments are designed keeping the following two main objectives in mind.

{\noindent\textbf{Objective 1:}} First of all, we will analyze how the proposed framework helps in getting better identification performance (in terms of recognition/identification accuracy) by choosing a sparse set of informative samples for annotation.
For this purpose we will compare the identification accuracy vs the number of images labeled, with the following baseline.
The baseline assumes that the representatives are chosen randomly for labeling and the classifier used for test is SRC.
Since, discriminative classifiers like SVM or random forests are required to train from scratch each time new samples are annotated, we do not use them in our online setting.

We also compared with a state-of-the-art representative selection framework - Sparse Modeling Representative Selection (SMRS)~\cite{Elhamifar2012a} which does not consider redundancy among chosen representatives.
Note that, in addition to the use of mid-level attributes (ref. section~\ref{sec:Relatedworks}), the experimental setting and performance metric (Cumulative Count Characteristic) in~\cite{Das2015} are also different.
The annotation effort in~\cite{Das2015} is measured in terms of how many binary match/non-match questions needs to be answered to get the images labeled.
The use of such performance metric in~\cite{Das2015} does not allow a direct comparison with the proposed method which employs a more traditional performance measure (recognition accuracy vs percentage of images labeled).

While comparing with the baselines shows the significance of informative representative selection over random selection for active labeling, the comparison with SMRS shows the role of redundancy reduction in the online setting.
For this purpose, we conducted experiments starting with unlabeled images with both balanced and imbalanced distributions of images per person.
Balanced and imbalanced scenarios are described in detail in section \ref{sec:sub:ward}.

{\noindent\textbf{Objective 2:}} The next objective is to study the scalability of the approach with a dataset containing a large number of persons.
The dataset considered here is an order of magnitude larger than in the above objective with respect to the number of people.
The performance measures and comparison baselines for this case are the same as in Objective 1.

There are lots of datasets that can be used to evaluate the proposed method for the said objectives.
However, some of these datasets (\textit{e.g.}, VIPER, GRID) contain too few images per person to form disjoint train and test sets in an online scenario.
We chose to experiment with three benchmark datasets - WARD~\cite{Martinel2012b}, iLIDS-VID~\cite{Wang2014}, and CAVIAR4REID~\cite{Cheng2011} as they contain a lot of persons with several images per person giving the opportunity to show performance improvement in an incremental manner as more and more unlabeled data are annotated.

{\noindent\textbf{Experimental Setup:}}
\begin{itemize}[noitemsep,nolistsep,leftmargin=*,topsep=0pt,partopsep=0pt]
\item Images have been normalized to $128\times 64$ to be consistent with the state-of-the-art person re-identification methods.
\item After segmenting the images into three salient regions (head, torso and legs)~\cite{Bazzani2012}, mean color feature (HSV) is generated following the scheme in~\cite{Hirzer2012}.
The head region is discarded, as it consists of a few and less informative pixels.
Each bodypart, is divided into blocks of size $8\times16$ and the blocks are overlapping by 50\% in horizontal and vertical directions.
\item The regularization parameter $\lambda_2$ is taken as $\lambda_0/\gamma$ where $\lambda_0$ is computed from the data~\cite{Elhamifar2012} and $\gamma$ is taken as 2.5 throughout.
For the other parameters, following values were used throughout, $\alpha = 0.2$ and $\beta = 0.3$.
For both WARD and CAVIAR4REID, $\lambda_1$ is taken as 2.
Since the number of people and images is more in iLIDS-VID, redundant examples are abundant compared to the other two datasets and thus $\lambda_1$ is taken as 10.
\item The dimension of the joint manifold has been taken as 10 throughout.
\item We ran all the experiments with 5 independent trials and report the average results. For each trial a unlabeled pool and a separate disjoint test set were created randomly.
\item As mentioned in section \ref{sec:RedundancyReductionIntra}, the threshold for intra-iteration redundancy reduction needs to be high so that  only very similar samples qualify as redundant images inside the hypergraph. The value was taken as 0.8 in the scale of similarity scores between 0 and 1. To compare fairly, the intra-iteration redundancy reduction step was applied to random selections too with same threshold value.
\item The proposed framework, generally, chooses different number of samples for labeling in each iteration for different unlabeled pool.
As a result, for different test sets, the accuracy may not be obtained for the same number of images labeled.
So we used spline interpolation to get the accuracies for the same number of labeled image.
For each experiment, the average accuracy vs labeled images plots were calculated taking the mean of this interpolated plots.
To show the robustness of the methods, we also show the corresponding $\pm$ standard deviation values of the accuracies too.
\end{itemize}

\subsection{WARD Dataset}
\label{sec:sub:ward}

The WARD dataset~\cite{Martinel2012b} has 4786 images of 70 different people acquired in a real surveillance scenario in 3 non-overlapping cameras.
It has large illumination variation along with resolution and pose changes.
This dataset is used to show the performance of the proposed framework starting with a balanced and an imbalanced pool of unlabeled images as two separate scenarios.
By `balanced' we mean that the pool is composed in such a way that each person has equal number of images per camera.
Though, in reality, such a perfectly balanced distribution of data is hard to come by, we conducted the experiments on such a balanced scenario to show that the proposed method performs well in such a scenario too.
For this dataset 2 random images per person per camera was chosen to form such a balanced pool.
The imbalanced pool was formed such that 20\% of the persons (\textit{i.e.}, 14) have 10 images, 50\% persons (\textit{i.e.}, 35) have 4 images and 30\% persons (\textit{i.e.}, 21) have 2 images per camera.
The test set for both the cases is composed of 2 images per person per camera.

\begin{figure*}[!t]
\centering
\subfigure[]{
\label{fig:ward_balanced}
\includegraphics[width=0.48\linewidth]{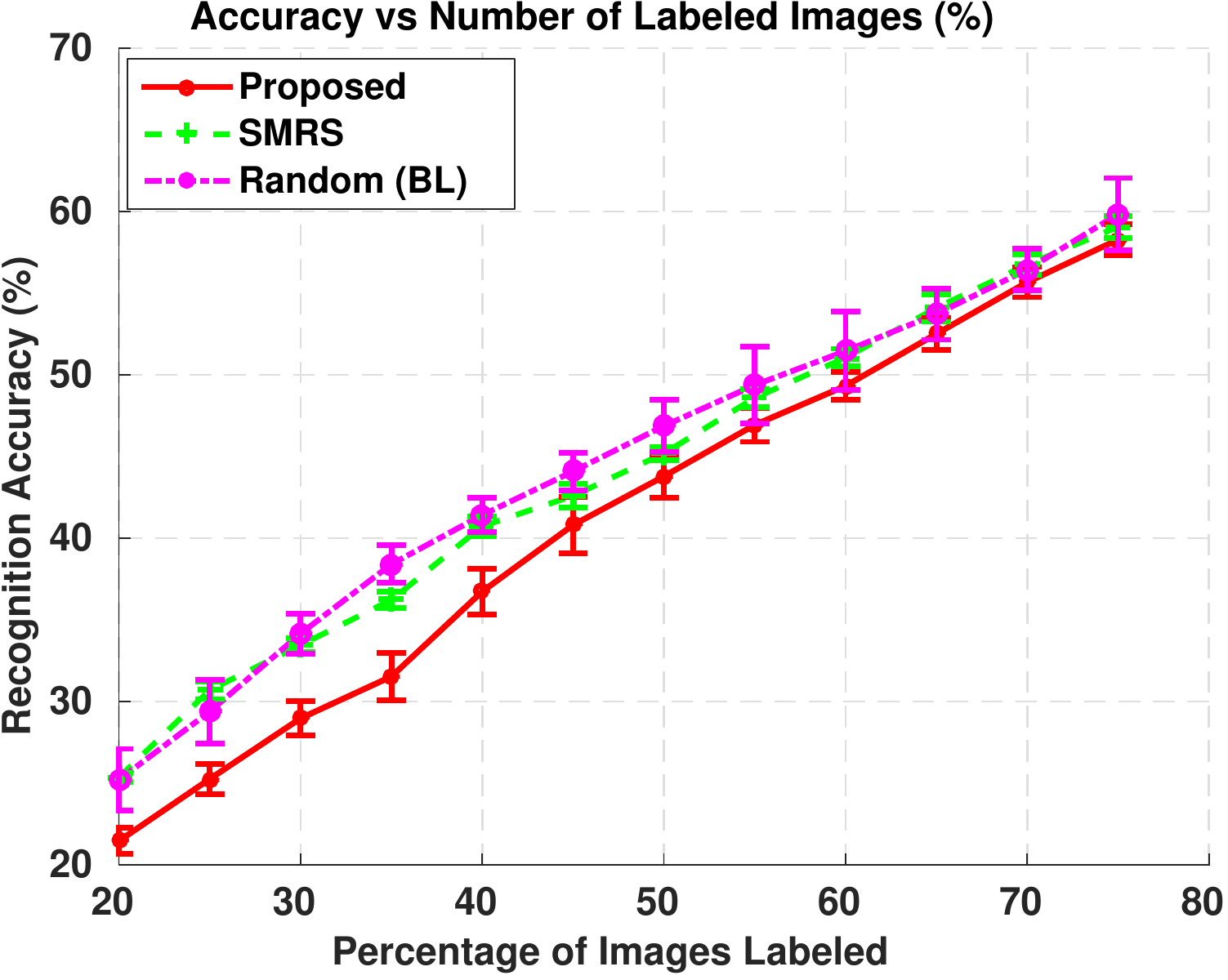}}
\subfigure[]{
\label{fig:ward_imbalanced_1}
\includegraphics[width=0.47\linewidth]{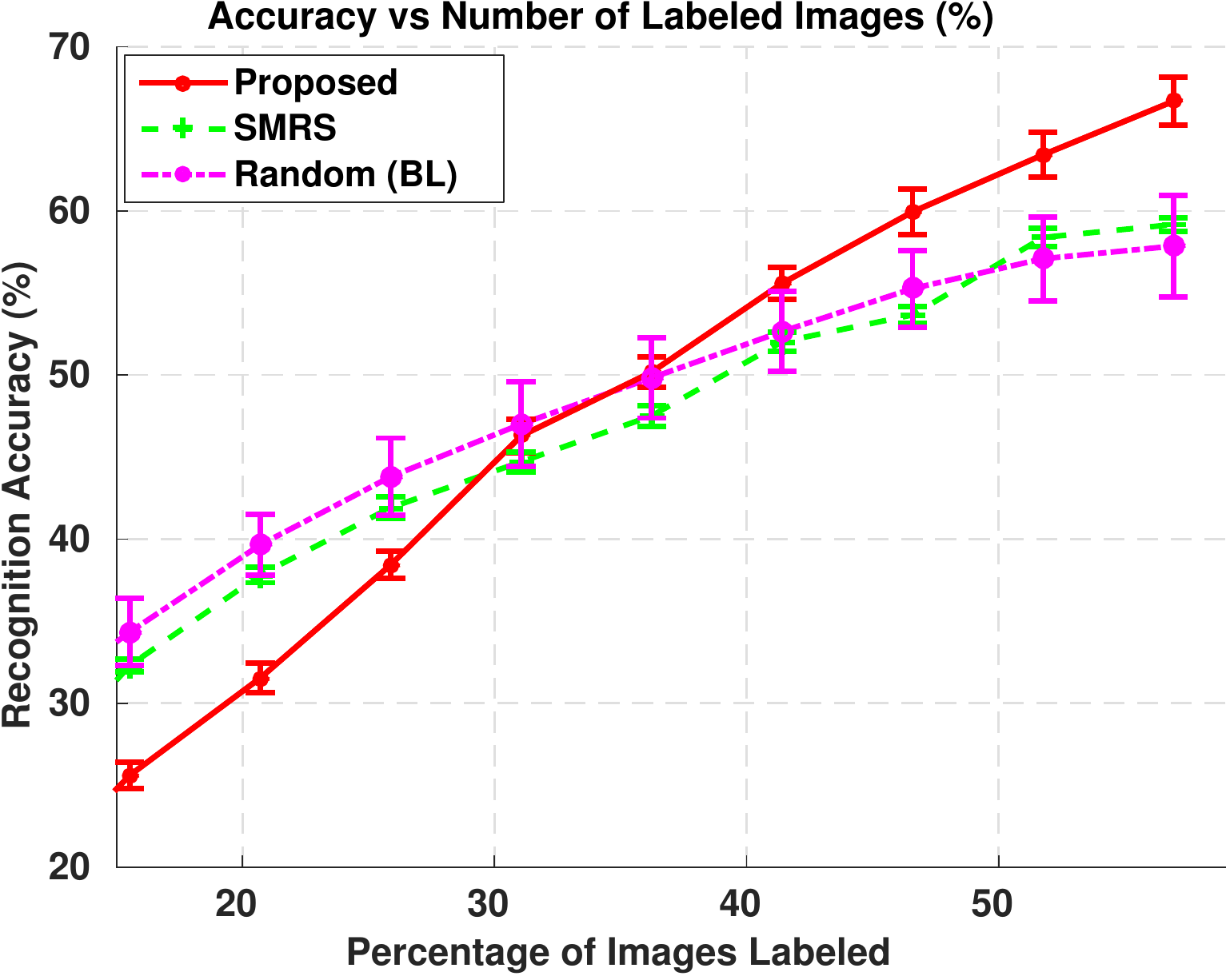}}
\caption
{Plot of testset accuracy (average) with the percentage of images labeled for the WARD dataset. Fig.~\subref{fig:ward_balanced}, \subref{fig:ward_imbalanced_1} show the performances for balanced and imbalanced set of unlabeled pools respectively.
}
\label{fig:ward}
\end{figure*}

Fig.~\ref{fig:ward_balanced} and \subref{fig:ward_imbalanced_1} show the comparative analysis of the test set accuracies as a function of the number of images labeled (as a percentage of the number of starting unlabeled images).
While the plots show the mean accuracy over 5 independent trials the vertical bars in each of the plot denote the corresponding standard deviation of the accuracy values around the mean.
In the balanced scenario, the number of images in the unlabeled pool to start with is 420 (70*2*3) and the accuracies are shown till around 70\% of the images are labeled.
For the imbalanced scenario the number of images in the starting pool is 966 and accuracies are shown till around 50\% of the images are labeled in Fig.~\ref{fig:ward_imbalanced_1}.
For random selection baseline in the balanced scenario, 21 random images (5\% of 420 unlabeled images) are chosen for annotation in each iteration.
In the imbalanced scenario, the number is 100 (10\% of 966 unlabeled images and then rounded to nearest greater multiple of 10).

{\noindent\textbf{Analysis of the Results:}} For both the balanced and the imbalanced data distribution, interesting observations can be made when performances of the three scenarios (viz. Baseline, SMRS and proposed method) are compared.
In the plots the acronym `BL' is used in place of `Baseline'.
For the balanced scenario, it can be seen that the three methods perform pretty closely.
Though SMRS and the baseline follow each other very closely, the baseline is more uncertain than both the representative selection based methods with or without considering redundancy.
This is shown by higher values of standard deviations for the baseline.
The superiority of the proposed method over SMRS can be observed in the more practical scenario when the data distribution is imbalanced (Fig.~\ref{fig:ward_imbalanced_1}).
Starting with lower recognition accuracy than both SMRS and the baseline, the proposed method surpasses SMRS when around 29\% images are annotated while it surpasses the baseline when around 35\% images are annotated.
With 57\% annotated data, the performance of the proposed method is better than the next best (SMRS) by around 7.5\%.

\begin{figure*}[t]
	\centering
	\includegraphics[width=1\linewidth]{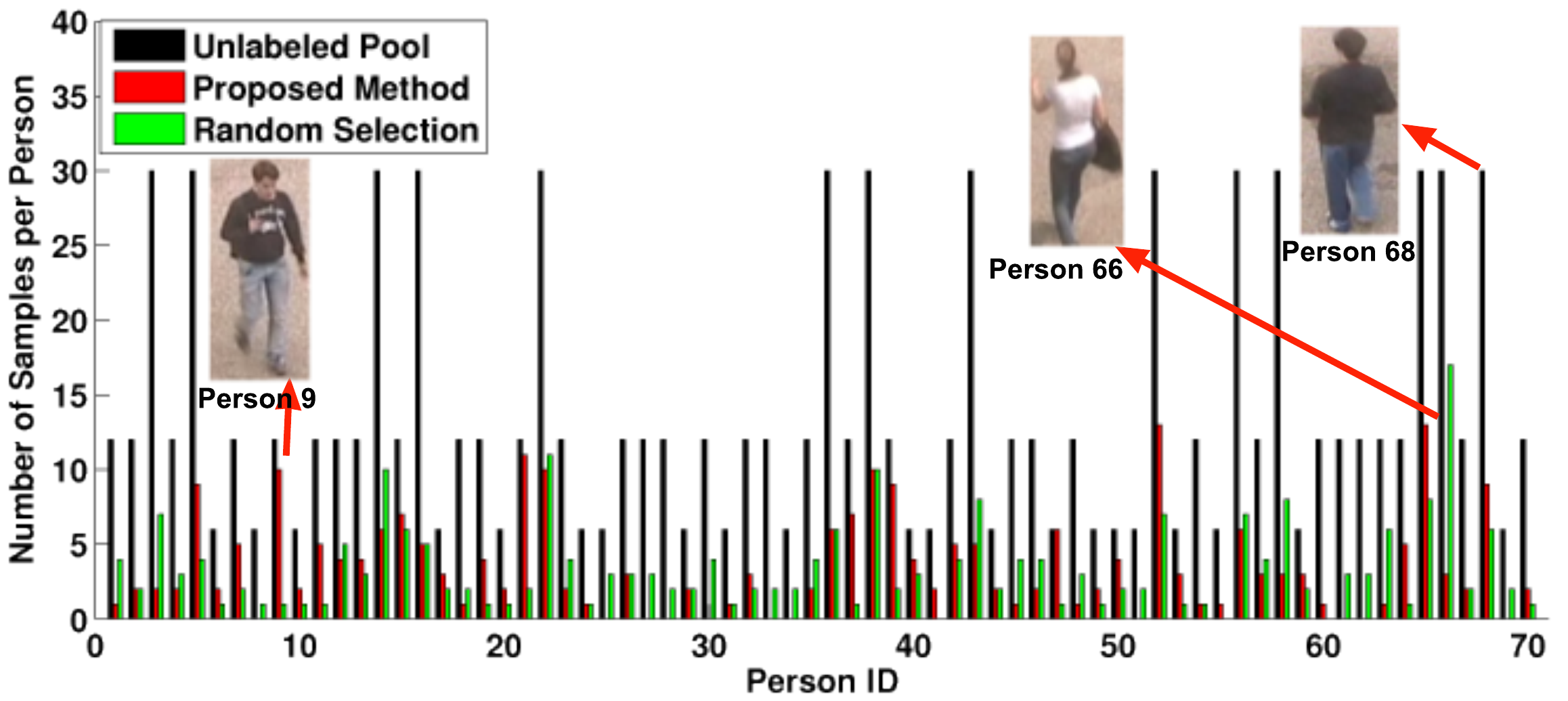}
	\caption{Imbalanced pool of unlabeled images. The three bars for each person (Id of the person is in the horizontal axis) give the number of images of that person in the starting unlabeled pool (black), in the annotated sets with proposed framework (red) and random selection (green). This snapshot is given after 25\% of the images in the imbalanced pool are labeled for each of the methods. See text (Sec.~\ref{sec:sub:ward}) for a detailed analysis of this figure.
	}
	\label{fig:ward_imbalanced_2}
\end{figure*}

When compared between the balanced and imbalanced scenarios, the uncertainty for all the methods are seen to be more for the imbalanced pool.
The imbalance in data distribution is, thus, seen to affect all the methods but the relatively large value of the error bars for the baseline method where random selection of images are made, shows that imbalance affects the proposed method less than it affects the random selection.
This is due to the reason that, in random representative selection, the samples are selected for annotation following a similar imbalanced distribution as the original pool.
On the other hand, the proposed method judiciously selects a diverse set of samples to negate the effect of imbalance in the data.
This can be seen more precisely in Fig.~\ref{fig:ward_imbalanced_2} where the three bars represent the number of samples per person in the starting imbalanced pool (black), in the annotated sets with proposed framework (red) and random selection (green) after 25\% of the unlabeled images are chosen by these two methods for labeling.
The horizontal axis shows the person IDs.
The distribution of images for annotation is seen to roughly follow the same distribution as that of the pool for the random selection while that is not the case for the proposed framework.
For example, person 68 and 9 look very similar and the proposed method chooses more number of images for both of them as they can create confusion than say, person 66 who looks markedly different.
This is done irrespective of the original distribution of the unlabeled pool.

\subsection{CAVIAR4REID Dataset}
\label{sec:sub:caviar4reid}

\begin{figure*}[!t]
\centering
\subfigure[]{
\label{fig:caviar_balanced}
\includegraphics[width=0.48\linewidth]{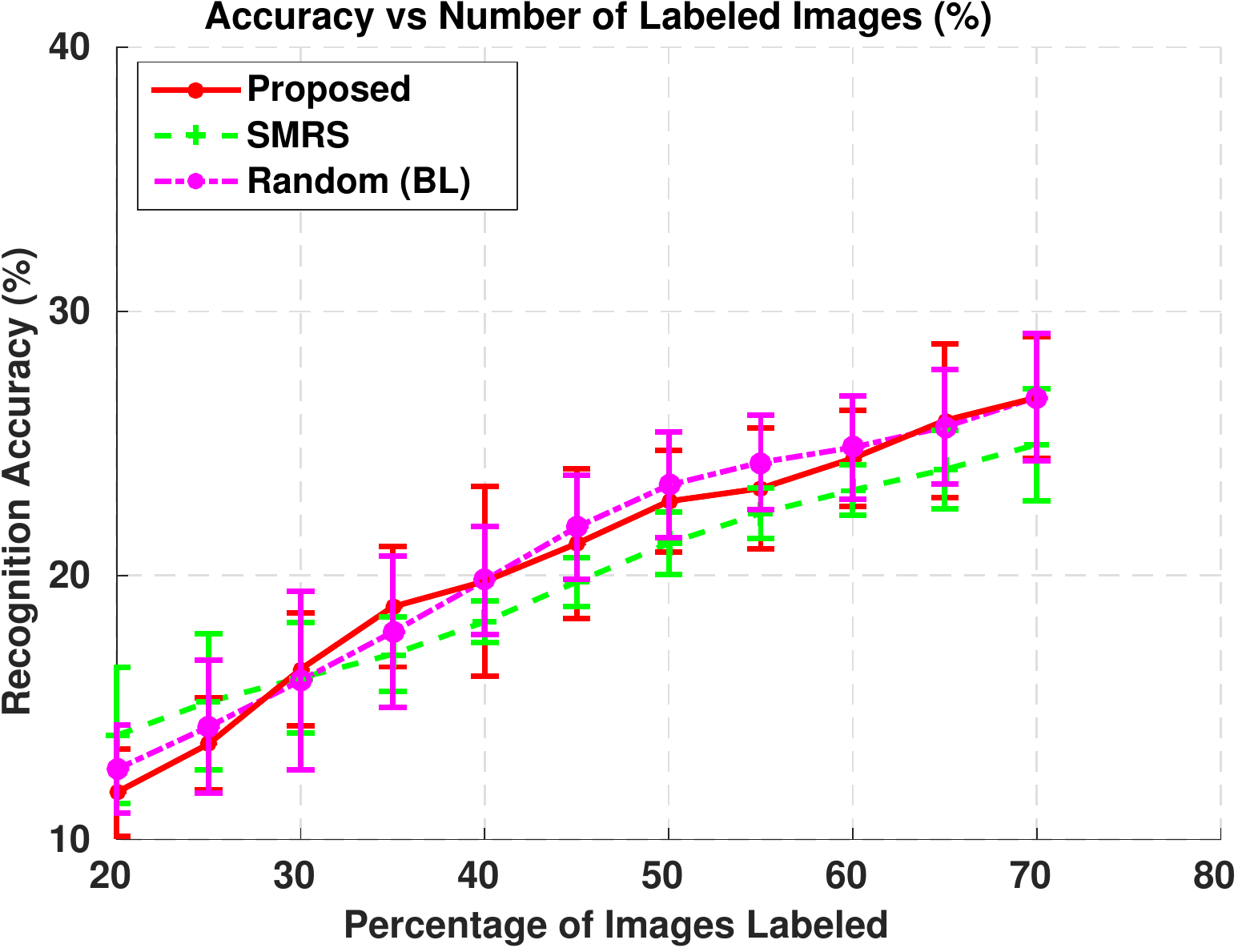}}
\subfigure[]{
\label{fig:caviar_imbalanced}
\includegraphics[width=0.47\linewidth]{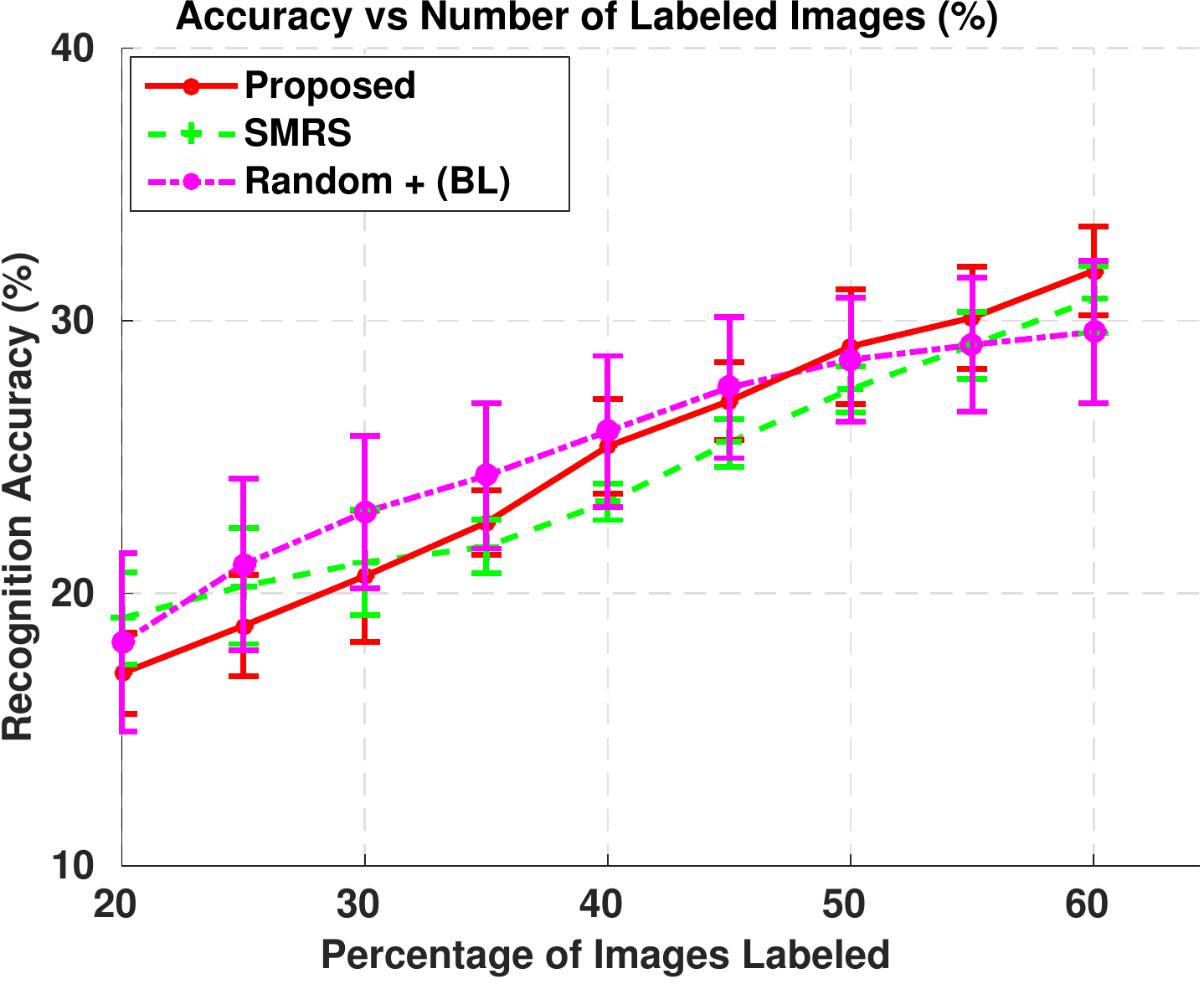}}
\caption
{Plot of testset accuracy (average) with the percentage of images labeled for the CAVIAR4REID dataset. Fig.~\subref{fig:caviar_balanced}, \subref{fig:caviar_imbalanced} show the performances for balanced and imbalanced set of unlabeled pools respectively.
}
\label{fig:caviar}
\end{figure*}

This dataset~\cite{Cheng2011} contains images of pedestrians extracted from the CAVIAR repository.
It is composed of images of 50 pedestrians viewed by two disjoint cameras.
The challenges in this dataset involve a broad change in the image resolution from $17\times39$ to $72\times144$ with severe pose variations, illumination changes and occlusion.
In terms of number of persons and cameras the dataset is of the scale as the WARD dataset.
The balanced pool, the imbalanced pool and the test pool are composed exactly same as done in WARD dataset.
This means that the starting pool of unlabeled images consists of 200 $(50*2*2)$ and 460 $(10*10*2 + 25*4*2 + 15*2*2)$
Fig.~\ref{fig:caviar_balanced} and \subref{fig:caviar_imbalanced} show the comparative analysis of the test set accuracies for this dataset as a function of the number of images labeled for the balanced and imbalanced set of unlabeled pools.

{\noindent\textbf{Analysis of the Results:}} For both the scenarios, the trend is similar to WARD.
For the balanced scenario, SMRS starts at the top but as new data arrives and thus the chance of redundant images becomes more plausible, the performance falls behind the other two.
The random baseline and the proposed method perform quite similarly reaching 26.75\% when 70\% of the unlabeled pool is annotated which is around 1.79\% higher than the SMRS at the same point. Similar to the WARD dataset, the proficiency of the proposed method is pronounced in the more practical imbalanced scenario.
The recognition accuracy is the highest among the three reaching a value of 31.83\% compared to 30.78\% and 29.59\% for SMRS and the random baseline respectively when 60\% of the unlabeled pool is annotated.

\begin{figure*}[!t]
\centering
\subfigure[]{
\label{fig:i-lids-vid_balanced}
\includegraphics[width=0.48\linewidth]{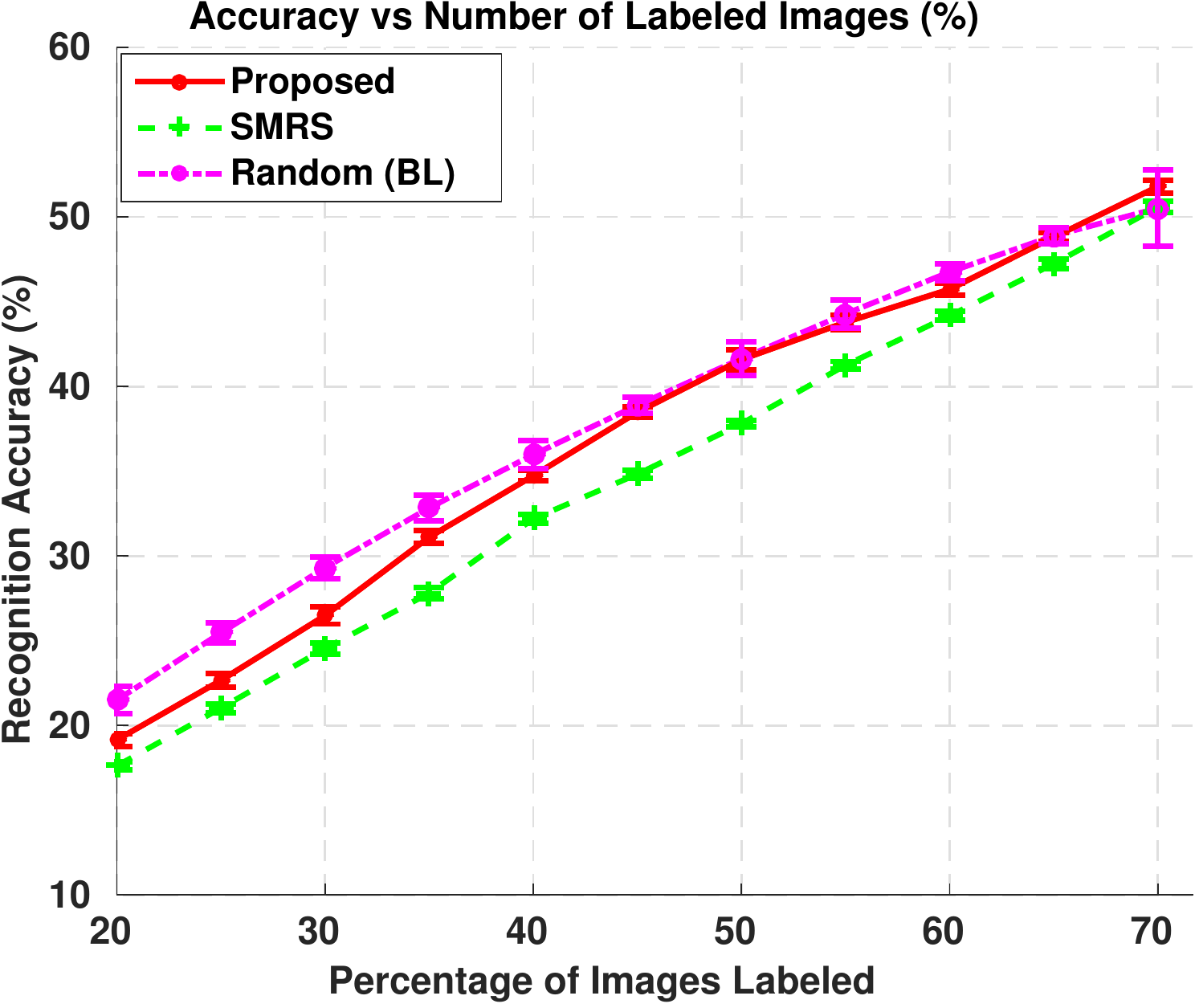}}
\subfigure[]{
\label{fig:i-lids-vid_imbalanced}
\includegraphics[width=0.48\linewidth]{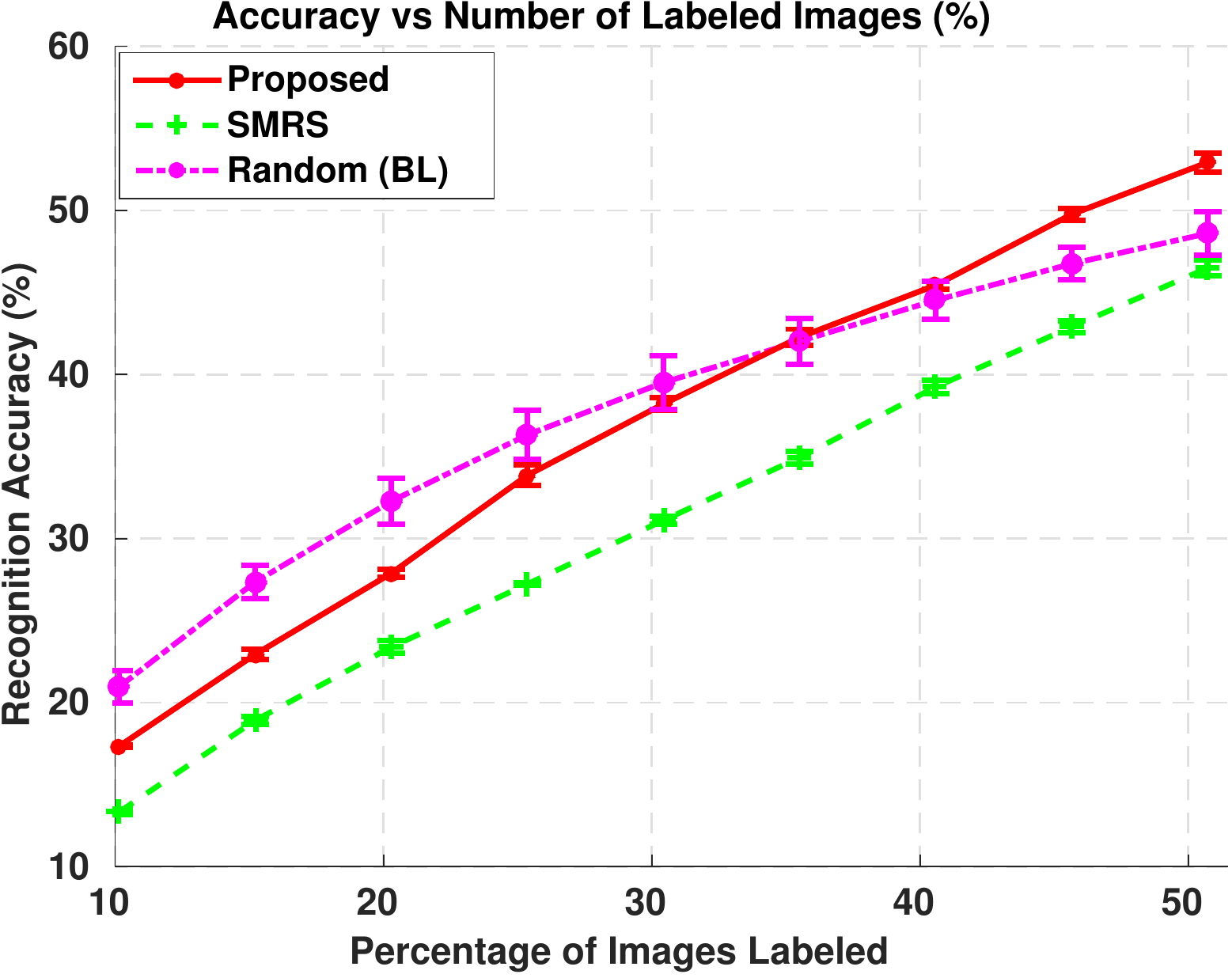}}
\caption
{Plot of testset accuracy (average) with the number of images labeled for the i-LIDS-VID dataset. Fig.~\subref{fig:i-lids-vid_balanced}, \subref{fig:i-lids-vid_imbalanced} show the performances for balanced and imbalanced set of unlabeled pools respectively.
}
\vspace{-8pt}
\label{fig:i-lids-vid}
\end{figure*}

\subsection{i-LIDS-VID Dataset}
\label{sec:sub:i_lids_vids}

iLIDS-VID~\cite{Wang2014} is a recently introduced person re-identification dataset.
This dataset consists of images from 300 people at an airport arrival hall captured through 2 non-overlapping cameras.
Apart from the typical challenges in person re-identification \textit{e.g.}, clothing similarities, clutter, lighting variations \textit{etc.}, one significant challenge in this dataset is the large number of people to be re-identified.
Following the same convention with the previous two datasets, here also we experiment in two scenarios - balanced and imbalanced data distributions.
The composition of unlabeled pool for both the scenarios are exactly same as the WARD or the CAVIAR4REID dataset.
However, due to the presence of more number of persons, the number of unlabeled images are much more than both of them.
For example, the numbers are 1200 for the balanced scenario and 2760 for the imbalanced scenario compared to 420 and 966 respectively for WARD.
The test set for this dataset also is composed of 2 images per person per camera.
Fig.~\ref{fig:i-lids-vid_balanced} and \subref{fig:i-lids-vid_imbalanced} show the comparative analysis of the test set accuracies for this dataset as a function of the number of images labeled for the balanced and imbalanced set of unlabeled pools.
The accuracies are shown till 70\% of the images in the unlabeled pool are labeled for the balanced scenario while for the imbalanced scenario accuracies are shown till around 50\% of the images in the unlabeled pool are labeled.
For random selection baseline in the balanced scenario, 120 random images (10\% of 1200 unlabeled images ) are chosen for annotation in each iteration.
In the imbalanced scenario, the number is 280 (10\% of 2760 and then rounded to nearest greater multiple of 10).

{\noindent\textbf{Analysis of the Results:}} For all the three methods, the trend is similar to that seen in WARD or CAVIAR4REID.
When compared to the results of these two datasets, for both the scenarios, the accuracies with same percentage of  labeled images are less for all these three methods.
For example the accuracies of the proposed method, SMRS and the baseline in the balanced scenario are 51.79\%, 50.58\%  and 50.52\% respectively for the i-LIDS-VID dataset with 70\% labeled images compared to 55.67\%, 56.75\% and 56.44\% for the WARD dataset and 26.75\%, 24.96\% and 26.75\% for the CAVIAR4REID dataset at the same percentage of labeled images.
This is due to the increased variability present in the i-LIDS-VID dataset with increased number of persons.
Fig.~\ref{fig:i-lids-vid_imbalanced} shows the comparison for the imbalanced scenario involving a lot more diverse and repetitive data.
The proposed method always performs better than the SMRS in this difficult dataset in this challenging scenario.
Starting with lower recognition accuracy than the baseline, the proposed method surpasses the baseline when around 35.5\% images are annotated.
With 50.72\% annotated data, the performance of the proposed method is better than the next best (baseline) by around 4.33\%.

For the imbalanced data distribution, the uncertainty is more in case of the random selection baseline than both the  representative selection based strategies (proposed and SMRS) with or without considering redundancy.
This is seen by the larger value of the standard deviations for the baseline.
For the case when the data distribution is balanced, the proposed method is at par to all the above methods.
Thus considering both the measures of performance (average as a measure of central tendency and standard deviation as a measure of variation) for the more practical scenario of imbalanced data, the proposed method is more robust than all the baselines as well as the representative selection based method but with no redundancy reduction term.

With the introduction of data-driven deep frameworks~\cite{Zhang2016,Wang2016}, it becomes easier to see that for the same algorithm, a dataset which has access to more annotations gives better performance.
For example, in~\cite{Zhang2016,Wang2016} we see a performance boost ranging from $10\%-30\%$ when the same method is applied to datasets (CUHK01~\cite{Li2013}, CUHK03~\cite{Li2014}, Market1501~\cite{Zheng2015}) where there are far more training images compared to datasets with lesser number of training images (VIPeR~\cite{Gray2007}).
The humongous amount of human involvement in annotating the training data can make such data-driven approaches impractical if efforts are not made to reduce the human effort without compromising the system performance much.
Instead of annotating blindly a large number of persons, our approach helps in the cause by judiciously focusing on a few informative samples to annotate.
Our experiments show that such a policy gets better performance at the cost of less annotation.

\subsection{Discussion about the Results}
\label{sec:sub:discussion}
Using the WARD (section~\ref{sec:sub:ward}), CAVIAR4REID (section~\ref{sec:sub:caviar4reid}) and i-LIDS-VID (section~\ref{sec:sub:i_lids_vids}) datasets, we have shown that the proposed method needs less samples to be labeled to attain same accuracy.
The effect is more pronounced in the imbalanced scenario where redundant samples are plenty.
This is clearly seen more towards the right in Fig.~\ref{fig:ward_imbalanced_1}, Fig.~\ref{fig:caviar_imbalanced} and~\ref{fig:i-lids-vid_imbalanced} where the proposed method outperforms the others when it matters the most \emph{i.e.}, when after a few iterations redundant samples come more frequently.
In i-LIDS-VID, the number of persons is comparatively more than the WARD or the CAVIAR4REID dataset which makes it easy to show the need for the redundancy reduction strategy as SMRS (where the redundancy reduction module is not there) is consistently below the proposed method.
The reason is that with more people comes more variability and thus redundancy in images may appear even if less number of images are labeled.
Thus two very different scales of these datasets helped us to show the effect of redundancy with more people as well as more labels.

\section{Conclusions}
\label{sec:conclusions}
In this work, we addressed the problem of creating a gallery of persons in an active learning set up with two different goals - reducing the labeling effort in presence of huge inflow of data and updating the model continuously so that it becomes adaptive to the changing dynamics of the data.
In doing so, a convex optimization based framework is proposed that progressively and judiciously chooses sparse and non-redundant set of samples for labeling.
A SRC classifier is used for online updation of the model.
Experiments on three publicly available benchmark datasets are performed to validate the proposed approach.
The future directions of our research will be to apply the framework to bigger networks with large numbers of cameras, and cope with wider horizons of computer vision \emph{e.g.}, online and continuous activity recognition.

\section{Acknowledgements} This work was partially supported by NSF grants CPS 1544969 and IIS-1316934.

\bibliography{ActiveLearningSparseReid}

\cleardoublepage
\begin{appendix}

\setcounter{secnumdepth}{0}
\section{Appendix}
This appendix/supplementary material accompanies the manuscript ``Continuous Adaptation of Multi-Camera Person Identification Models through Sparse Non-redundant Representative Selection''.
In this supplementary material we are providing the time complexity analysis of the FISTA[55] steps involved in finding the redundancy restricted representatives and the sparse representations of the test set.

\setcounter{secnumdepth}{1}
\section{Time Complexity Analysis}
The rate of convergence of FISTA algorithm in terms of the number of iterations is provided in~\cite{Beck2009}. Here, we are providing the following time complexity per FISTA step. FISTA being an iterative method, it is important to note the time complexity of the FISTA steps for each iteration. The major contributor to the time complexity for the representative selection phase are the matrix multiplications in computing the gradient and Lipschitz constants $\nabla g(X)$ and $L_g$ respectively (ref. eqn. (12) in the main paper). Considering the sizes of the 3 matrices $Z,\hat{Z}_{0}$ and $X$ as provided in Table 1 of the main paper, we can see that the time complexity of computing the gradient $\nabla g(X)$ is $O(n^{2}d) + O(n^{2}d+n^{3}) + O(ndn_{0}+n^2n_{0}+n^3)$ while the same for computing the Lipschitz constant $L_g$ is $O(n^2d+n^2) + O(ndn_{0}+n^{2}n_{0}+n^2)$. Since $d$ is a constant, we can further reduce the above complexities to $O(n^3)$ and $O(n^2)$ respectively.
Similarly, the SRC classifier also employs FISTA algorithm for the optimization. While other algorithms specifically tuned for sparse optimization (e.g., LASSO, LARS etc.) could have been used (which may have provided better time complexity depending on the dictionary element dimension and the number of dictionary elements), we opted for FISTA as we used the same algorithm for representative selection and thus makes the whole framework more general. 
The major contributor to the time complexity for the SRC classification are the matrix multiplications in computing the gradient and Lipschitz constants $\nabla p(C)$ and $L_p$ respectively (ref. eqn. (13) of the main paper). Assuming the number of test images to be $N$ (i.e., the sizes of the matrices $\hat{Y}_{0},Y,C$ and $L$ to be $d_1 \times n_0$, $d_1 \times N$, $n_0 \times N$ and $N \times N$- in our case $d_1$ extracted feature dimension and $N$ is the number of test images), we can see that the time complexity of computing the gradient $\nabla p(C)$ is $O(n_{0}d_{1}n)+O(n_{0}^{2}d_1 + n_{0}^{2}N)+O(n_0N^2)$ while the same for computing the Lipschitz constant $L_p$ is $O(n_{0}^{2}d_1 + n_{0}^{2})+O(N^2)$. It should be noted that the computation of the proximal operators (ref. eqn. (14) and (15) of the main paper) also contributes to the complexity per FISTA iteration. However, these are second order only $(O(n^2)$ and $O(n_{0}N)$ respectively) and thus the major contributor to the computational complexity are as given earlier.

\end{appendix}

\end{document}